\documentclass{magicstyle/magic}

\usepackage{amsmath,amssymb,amsfonts}
\usepackage{xspace}
\usepackage{enumitem}
\usepackage{makecell}
\usepackage{pifont}
\usepackage{todonotes}
\setlist[itemize]{leftmargin=*,itemsep=0em,parsep=0.3em,topsep=0.3em}
\usepackage{array}
\usepackage{colortbl}
\usepackage{tabularx}
\usepackage{algorithm}
\usepackage{algpseudocode}

\makeatletter
\newcommand{\tablebf}[1]{{\fontsize{\dimexpr\f@size pt+0.5pt\relax}{\dimexpr\f@size pt+3pt\relax}\selectfont\bfseries\sffamily #1}}
\makeatother
\AtBeginEnvironment{tabular}{\let\textbf\tablebf}
\AtBeginEnvironment{tabularx}{\let\textbf\tablebf}

\newcommand{\oursrow}{\rowcolor[gray]{0.92}}

\let\cite\citep

\newsavebox{\fitwidthbox}
\newcommand{\fitwidth}[1]{\sbox{\fitwidthbox}{#1}%
  \ifdim\wd\fitwidthbox>\linewidth\resizebox{\linewidth}{!}{\usebox{\fitwidthbox}}\else\usebox{\fitwidthbox}\fi}

\newcommand{\github}{\raisebox{-1.5pt}{\includegraphics[height=1.05em]{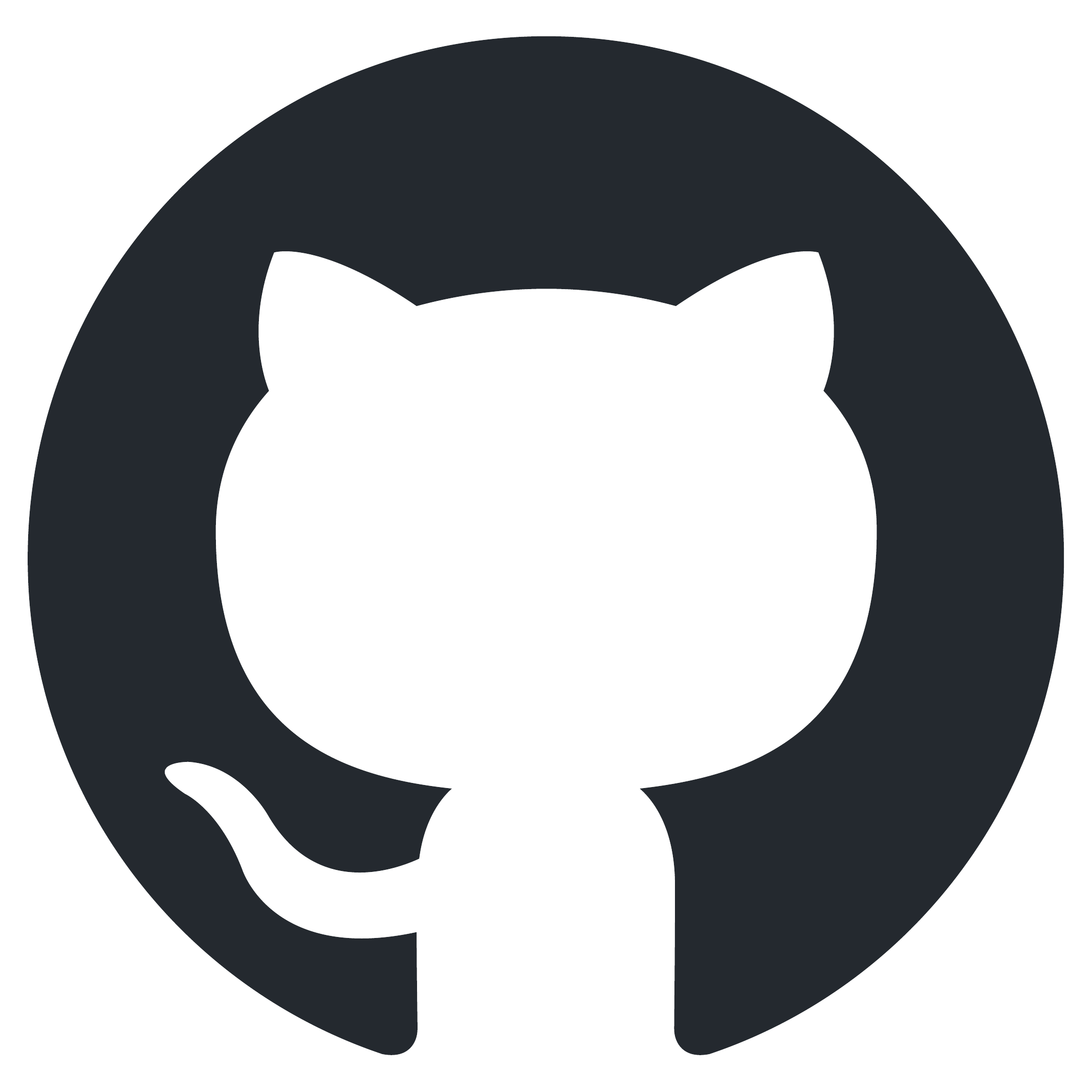}}\xspace}

\newcommand{\emailLogo}{\raisebox{-1.5pt}{\includegraphics[height=1.05em]{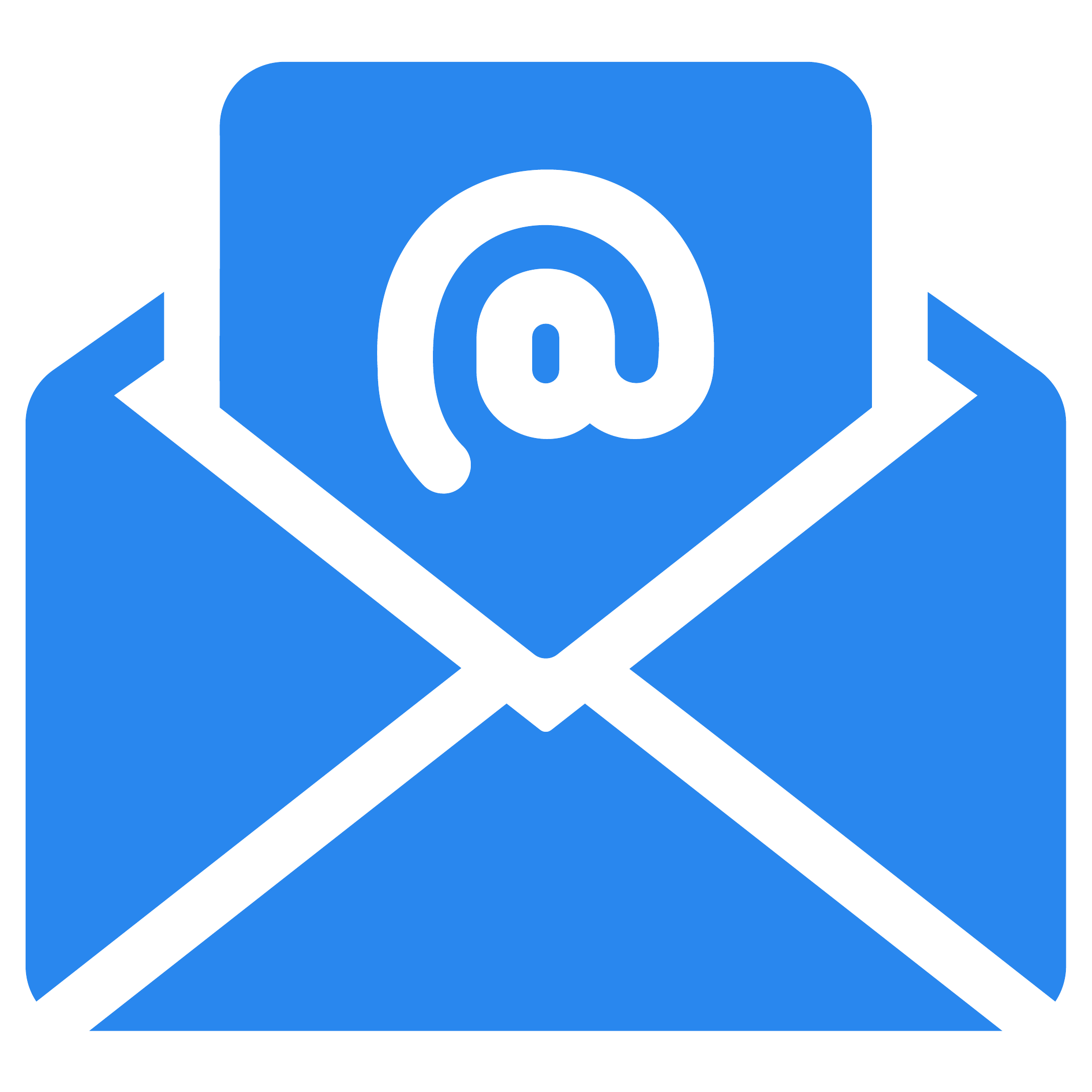}}\xspace}
\newcommand{\websiteLogo}{\raisebox{-1.5pt}{\includegraphics[height=1.05em]{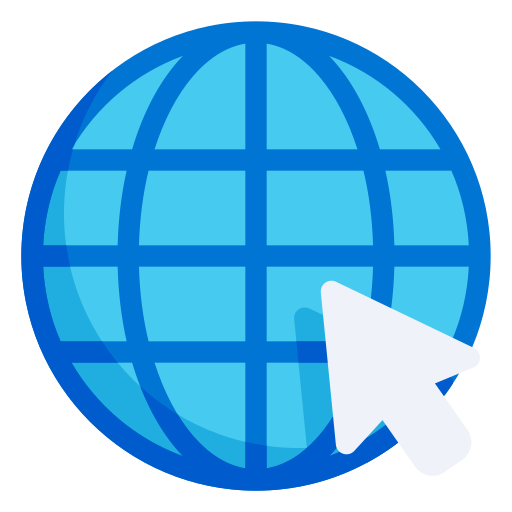}}\xspace}

\title{\textcolor{magicblue}{BiCAT: Bimanual Coordination Action Tokenization}}

\author[1]{Xuanwei Liu}
\author[1]{Guipu Cao}
\author[1]{Zhongyi Luo}
\author[1]{Jiafei Duan*}

\affiliation[1]{National University of Singapore}

\contribution[*]{Corresponding Author}

\abstract{Autoregressive vision-language-action (VLA) models predict discrete action tokens with the next-token objective of their vision-language backbone, so the action tokenizer defines the interface between the policy and the robot. In bimanual manipulation, the two arms share a repertoire of movements yet coordinate through complementary roles, and an action representation should capture both. However, most existing tokenizers encode both arms as a single concatenated action vector, in which motion patterns common to the two arms are not shared and cross-arm dependencies remain entangled within each token. The policy must therefore learn both aspects implicitly. We propose Bimanual Coordination Action Tokenization (BiCAT), which shares one motion representation across arms while conditioning each arm’s tokens on its partner through symmetric cross-arm attention and a joint decoder. The policy then predicts all tokens of each arm in parallel, generating the left arm first and the right arm conditioned on it. Across ten RoboTwin 2.0 tasks and six real-world tasks on three embodiments, BiCAT achieves the highest success among the compared tokenizers. In simulation, it improves mean success by 17.6 percentage points over the strongest of eight baselines, with the largest gains on tasks that require close coordination between the arms, while requiring only 40.5 ms per action chunk. These results suggest that the action tokenizer is a natural place to model bimanual coordination.

}

\metadata[\websiteLogo Webpage:]{\href{https://magiclab-nus.github.io/BiCAT/}{\texttt{magiclab-nus.github.io/BiCAT}}}

\metadata[\vspace{.3em}\github Code:]{\href{https://github.com/MAGICLAB-NUS/BiCAT}{\texttt{github.com/MAGICLAB-NUS/BiCAT}}}

\metadata[\vspace{.3em}\emailLogo Contact:]{\email{duanj1@nus.edu.sg}}

\begin{document}

\maketitle

\begin{figure}[!ht]
    \centering
    \includegraphics[width=\textwidth]{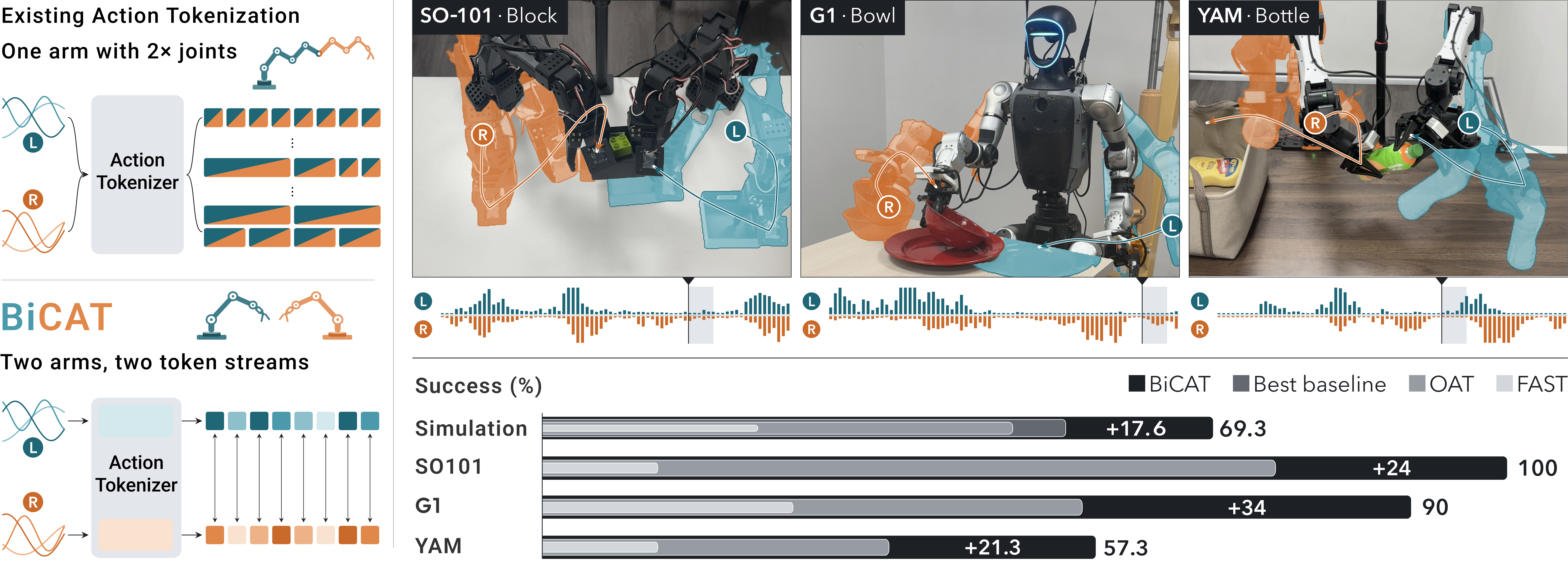}
    \caption{\textbf{A bimanual robot is two arms, not one arm with
    twice the joints.}
    Left: existing tokenizers treat the two arms as a single arm and mix the
    \textcolor[HTML]{1F6676}{\textbf{left}} and
    \textcolor[HTML]{C96A2B}{\textbf{right}} arm in every token; BiCAT gives
    each arm its own token stream, aware of its partner.
    Right: BiCAT rollouts on three embodiments with each arm's motion over
    time, and success rates in simulation and on real robots; white numbers
    give the gain over the best baseline.}
    \label{fig:teaser}
\end{figure}

\section{Introduction}
\label{sec:introduction}

Vision-language-action (VLA) models extend pretrained vision-language models to robot control~\cite{brohan2023rt2,kim2024openvla,qu2025spatialvla,black2024pi0,black2025pi05,bjorck2025gr00t,fang2026molmoact2}. A common design keeps the autoregressive next-token prediction of the backbone: an action tokenizer discretizes continuous action chunks into tokens, which the policy generates either one token at a time~\cite{brohan2022rt1,brohan2023rt2,kim2024openvla,pertsch2025fast,qu2025spatialvla} or in blocks of tokens predicted in parallel~\cite{liu2026oat_rss,jawaid2026nac,dong2026actioncodec,liu2026faster}. The action tokenizer thus determines what the policy learns to predict and how errors in those predictions affect the resulting robot motion~\cite{pertsch2025fast,liu2026oat_rss,dong2026actioncodec}. Existing tokenizers have explored frequency-domain compression~\cite{pertsch2025fast}, learned temporal abstraction~\cite{mete2024quest}, vector quantization~\cite{vandenoord2017vqvae,lee2024vqbet,wang2025vqvla}, and coarse-to-fine coding~\cite{liu2026oat_rss,jawaid2026nac}, but they treat an action chunk as a flat vector and do not model how its dimensions are organized across the arms of the robot.

This limitation is especially consequential in bimanual manipulation, where successful execution depends on coordinated motion across both arms. Tasks such as handing over objects, folding fabric, and carrying large objects often require two arms~\cite{smith2012dual,krebs2022bimanual}, and VLAs are increasingly being applied to these tasks~\cite{liu2024rdt,black2024pi0,bjorck2025gr00t,chen2025robotwin2}. However, most general-purpose action tokenizers concatenate both arms' actions into a single vector, a formulation we call \emph{joint tokenization}. Although this formulation can capture cross-arm dependencies, it leaves the structure of bimanual coordination to be learned implicitly.

Human bimanual manipulation illustrates two useful aspects of this structure: the arms share a repertoire of movements, yet adapt their actions to complementary roles~\cite{guiard1987asymmetric}. During a handover, for example, one hand presents and releases the object while the other reaches to receive and secure it. For robots with matching arm kinematics, this suggests an action representation that captures both reusable motion patterns across arms and task-dependent coordination between them. Joint tokenization does not explicitly distinguish these two aspects, leaving the policy to learn both shared motion structure and cross-arm dependencies from the concatenated representation.

Making this structure explicit raises a design question. One option, which we call \emph{per-arm tokenization}, splits the action by arm and tokenizes each arm separately, as in $\mathcal{M}^{2}$Tok~\cite{xu2026m2tok}, leaving coordination to the policy. Per-arm tokenization yields arm-indexed tokens and lets both arms reuse one motion representation, but it removes partner context from each arm's codes; in our experiments, such a tokenizer performs no better than the strongest joint tokenizer (Sec.~\ref{sec:simulation-control}). This paper therefore asks: \emph{once the arms are tokenized as separate streams, how should the streams interact to yield a better interface for bimanual policies?}

Our key insight is that arm-indexed tokens need not be arm-independent: each arm's tokens can be computed with access to its partner while the two streams remain separate. We build on this insight in \textbf{Bimanual Coordination Action Tokenization (BiCAT)}, a per-arm tokenizer that restores partner context. Its design follows one principle: \emph{share the motion representation across arms while conditioning discretization on the partner arm}. For the shared motion patterns, \textbf{BiCAT} exploits the mirror-image mounting of the arms: it maps the right arm into the left arm's convention, so that mirrored motions share one canonical trajectory, and encodes both arms with one shared causal temporal encoder. For the cross-arm coordination, symmetric cross-arm attention adds partner context to each stream before shared finite scalar quantization (FSQ)~\cite{mentzer2024fsq}, and a joint decoder reconstructs both trajectories from temporally paired codes. Trained by reconstruction alone, \textbf{BiCAT} produces separate tokens for each arm, which the VLA predicts in two successive blocks: the tokens of each arm are generated in parallel, and the right arm is conditioned on the left.

We compare \textbf{BiCAT} with eight baselines from seven tokenizer families under a common VLA training protocol. On ten RoboTwin 2.0 tasks~\cite{chen2025robotwin2}, \textbf{BiCAT} improves mean success by \textbf{17.6} percentage points over the strongest baseline, with a gain that grows from 1.5 points on single-arm tasks to 21.3 on tightly coupled ones, and on six real-world tasks across three embodiments it succeeds in 113 of 150 trials, compared with 74 for OAT~\cite{liu2026oat_rss}. Two-block prediction keeps policy latency at least $5.6\times$ lower than token-by-token decoding. Ablations show that the gain requires both the tokenizer and the prediction schedule: removing cross-arm interaction or predicting both arms in one block leaves \textbf{BiCAT} no better than the strongest baseline.

Our contributions are threefold:
\begin{itemize}
    \item \textbf{BiCAT}, a per-arm action tokenizer that shares one motion representation across mirrored arms and exchanges partner context on both sides of the discrete bottleneck.
    \item Improved bimanual VLA control over eight baselines from seven tokenizer families in simulation, and over FAST and OAT on three real embodiments.
    \item An analysis of per-arm tokenization showing that cross-arm interaction inside the tokenizer and arm-wise block prediction in the policy are jointly required for the gain over joint tokenization.
\end{itemize}

\section{Related Work}
\label{sec:related}

\subsection{Vision-Language-Action Models}
\label{sec:related-vla}
VLAs build on vision-language models pretrained on internet-scale image and text data~\cite{beyer2024paligemma,steiner2024paligemma2} and are fine-tuned on large robot datasets~\cite{oxe2024,khazatsky2024droid} for control~\cite{driess2023palme,brohan2023rt2,kim2024openvla,black2024pi0,lee2025molmoact,fang2026molmoact2}, inheriting the instruction following and generalization of their backbones~\cite{brohan2023rt2,kim2024openvla,black2025pi05}. Existing VLAs differ mainly in how they output actions. Continuous approaches regress actions from the backbone or attach a generative action expert, such as a diffusion or flow-matching model~\cite{kim2025oft,black2024pi0,black2025pi05,bjorck2025gr00t,shukor2025smolvla}. Discrete approaches tokenize actions and keep the autoregressive next-token objective of the backbone~\cite{brohan2023rt2,kim2024openvla,pertsch2025fast,qu2025spatialvla,wang2025vqvla,goyal2025vla0,liu2026faster}. Discrete prediction can incur higher inference latency~\cite{pertsch2025fast,liu2026faster}, but it trains the backbone with the same next-token objective as its pretraining, which better preserves its pretrained knowledge than propagating a continuous action-expert loss into it~\cite{driess2025knowledge}. Both the latency and the learnability of discrete prediction depend on the action tokenizer~\cite{pertsch2025fast,liu2026oat_rss,dong2026actioncodec}, which makes it a central design choice for VLAs; we study it in the bimanual setting.

\subsection{Action Tokenization}
\label{sec:related-tokenization}
Early discrete-action VLAs such as RT-1, RT-2, and OpenVLA~\cite{brohan2022rt1,brohan2023rt2,kim2024openvla} discretize each action dimension into bins and predict the resulting symbols autoregressively, which yields long sequences as the action dimension and horizon grow. FAST~\cite{pertsch2025fast} shortens them with a discrete cosine transform (DCT) and byte-pair encoding (BPE), at the cost of variable-length sequences that can decode into invalid actions~\cite{liu2026oat_rss}. Learned tokenizers instead compress action chunks into fixed sets of latent symbols: VQ-BeT~\cite{lee2024vqbet} and VQ-VLA~\cite{wang2025vqvla} use vector quantization, QueST~\cite{mete2024quest} combines causal temporal encoding with finite scalar quantization~\cite{mentzer2024fsq}, and OAT~\cite{liu2026oat_rss}, NAC~\cite{jawaid2026nac}, and CATok~\cite{zhang2026catok} order their tokens from coarse to fine.

Beyond compression and reconstruction fidelity, recent work has examined how action representations affect downstream policy learning. ActionCodec~\cite{dong2026actioncodec} investigates tokenizer properties associated with VLA optimization, while X-Tokenizer~\cite{kang2026xtokenizer} and ActionPiece~\cite{lian2026actionpiece} incorporate vision-language alignment and relational supervision between action chunks, respectively. When applied to bimanual robots, most existing discrete tokenizers operate on the concatenated action vector of both arms, the joint tokenization of Sec.~\ref{sec:introduction}. $\mathcal{M}^{2}$Tok~\cite{xu2026m2tok} expands quantization capacity through multiple codebooks and tokenizes each arm separately, but does not model cross-arm interaction within the tokenizer, leaving its role in discrete action representations largely unexplored.

\subsection{Bimanual Action Representations}
\label{sec:related-bimanual}
Bimanual policies typically represent both arms as one concatenated action vector. ACT~\cite{zhao2023act} and ALOHA Unleashed~\cite{zhao2024alohaunleashed} learn bimanual skills through action-chunk prediction, RDT-1B~\cite{liu2024rdt} develops a large-scale diffusion policy for bimanual manipulation, and general-purpose VLAs have been applied to bimanual robots~\cite{black2024pi0,bjorck2025gr00t} and evaluated on bimanual manipulation benchmarks~\cite{chen2025robotwin2,wang2026roboeval}. Explicit inter-arm structure has mainly been introduced in the policy: InterACT~\cite{lee2024interact} adds synchronization blocks between arm decoders. In the tokenizer, $\mathcal{M}^{2}$Tok~\cite{xu2026m2tok} separates the arms but keeps them independent. \textbf{BiCAT} instead models cross-arm interaction within a learned discrete action tokenizer, producing arm-indexed tokens through cross-arm attention and joint reconstruction.

\section{Method}
\label{sec:method}

\begin{figure*}[t]
    \centering
    \includegraphics[width=\textwidth]{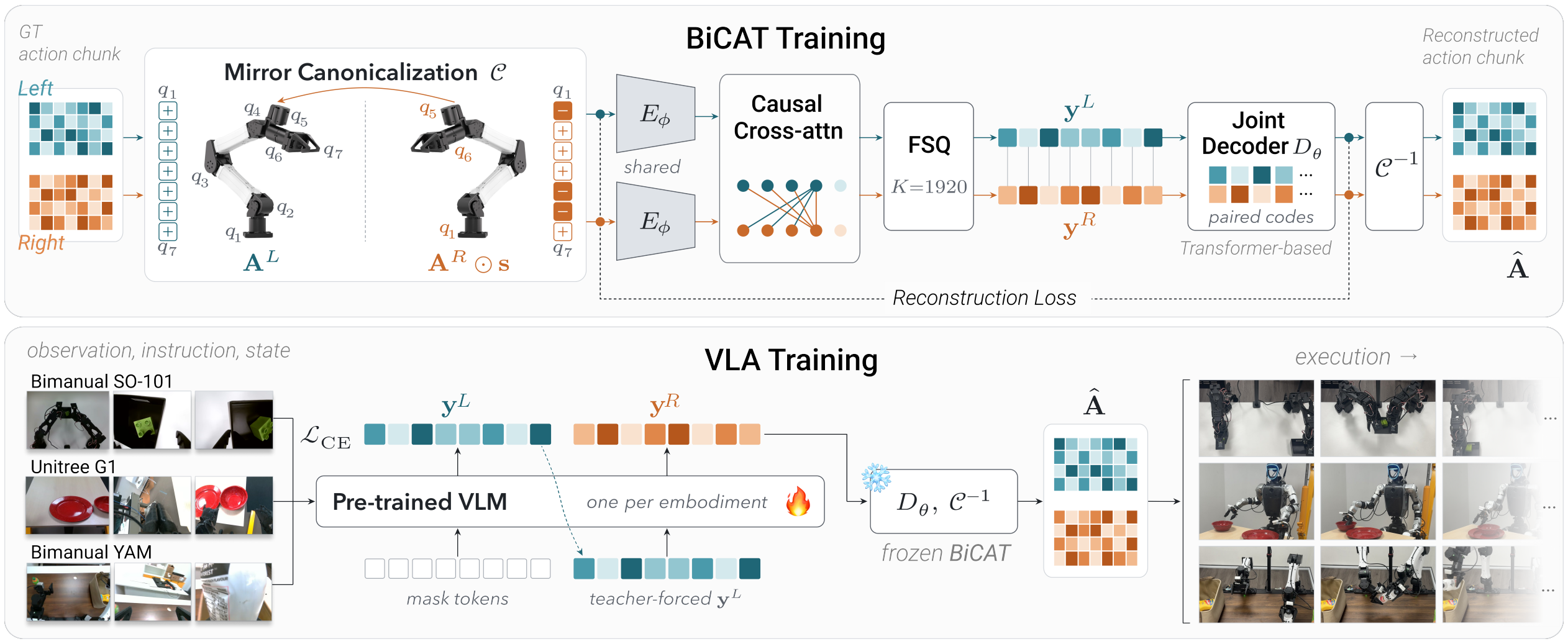}
    \caption{Overview of BiCAT. Top: mirror canonicalization $\mathcal{C}$
    flips the signed joints of the right arm ($q_1,\ldots,q_7$, with $q_7$ the
    gripper) and applies shared normalization; the shared encoder $E_{\phi}$
    and causal cross-attention give each arm its partner's context before FSQ;
    and the joint decoder $D_{\theta}$ reconstructs the canonical chunk under
    $\mathcal{L}_{\mathrm{rec}}$, which $\mathcal{C}^{-1}$ maps back to
    $\widehat{\mathbf{A}}$. Bottom: a VLM predicts $\mathbf{y}^{L}$, then
    $\mathbf{y}^{R}$; the frozen decoder returns actions, executed on
    Bimanual SO-101, Unitree G1, and Bimanual YAM. Teal: left arm; orange:
    right arm.}
    \label{fig:bicat_overview}
\end{figure*}

\textbf{BiCAT} tokenizes a bimanual action chunk into two arm-indexed token
streams that share one motion representation and are computed with access to
each other (Fig.~\ref{fig:bicat_overview}). Sec.~\ref{sec:preliminaries}
introduces action tokenization in the bimanual setting,
Sec.~\ref{sec:bicat_tokenization} describes the tokenizer, whose training is
summarized in Algorithm~\ref{alg:bicat_training}, and
Sec.~\ref{sec:bicat_policy} describes how a VLA predicts its tokens.

\subsection{Preliminaries}
\label{sec:preliminaries}

\paragraph{Action tokenization.}
A VLA policy maps an observation $\mathbf{o}$, comprising camera images,
robot state, and a language instruction, to a chunk of $H$ future actions
$\mathbf{A}\in\mathbb{R}^{H\times D}$. To retain the next-token prediction
of its backbone, an action tokenizer $\mathcal{T}$ maps the chunk to discrete
tokens $\mathbf{y}=\mathcal{T}(\mathbf{A})$ from a vocabulary of size $K$,
and a detokenizer $\mathcal{T}^{-1}$ maps tokens back to an executable chunk
$\widehat{\mathbf{A}}=\mathcal{T}^{-1}(\mathbf{y})$. Grouping the tokens into
$B$ blocks $\mathbf{y}_{1},\ldots,\mathbf{y}_{B}$, the policy with
parameters $\theta_{\mathrm{V}}$ is trained on demonstrations $\mathcal{D}$
to maximize
\begin{equation}
    \mathbb{E}_{(\mathbf{o},\mathbf{A})\sim\mathcal{D}}
    \Big[\sum_{b=1}^{B}\log p_{\theta_{\mathrm{V}}}
    (\mathbf{y}_{b}\mid\mathbf{o},\mathbf{y}_{<b})\Big],
    \label{eq:ar_objective}
\end{equation}
where the tokens within a block are predicted in parallel. Token-wise
autoregression uses blocks of one token~\cite{kim2024openvla,pertsch2025fast},
and block-wise prediction~\cite{stern2018blockwise,liu2026oat_rss,jawaid2026nac,dong2026actioncodec,liu2026faster}
reduces the number of sequential decoding steps to $B$.

\paragraph{Bimanual action chunks.}
A bimanual chunk $\mathbf{A}=[\mathbf{A}^{L}\Vert\mathbf{A}^{R}]
\in\mathbb{R}^{H\times 2d_a}$ concatenates the actions of the left and right
arms, where $d_a$ is the per-arm action dimension and $\Vert$ denotes feature
concatenation. Joint tokenization applies one tokenizer to the whole chunk,
$\mathbf{y}=\mathcal{T}(\mathbf{A})$, so that a token can mix both arms.
Per-arm tokenization applies it to each arm separately,
$\mathbf{y}^{r}=\mathcal{T}(\mathbf{A}^{r})$ for $r\in\{L,R\}$, which yields
arm-indexed tokens but computes each arm's tokens without its partner. \textbf{BiCAT}
keeps arm-indexed token streams but computes them from both arms:
\begin{equation}
    (\mathbf{y}^{L},\mathbf{y}^{R})=\mathcal{T}(\mathbf{A}^{L},\mathbf{A}^{R}),
    \qquad
    \widehat{\mathbf{A}}=\mathcal{T}^{-1}
    (\mathbf{y}^{L},\mathbf{y}^{R}),
    \label{eq:bicat_mapping}
\end{equation}
where $\mathbf{y}^{r}\in\{0,\ldots,K-1\}^{M}$ contains $M$ tokens for arm
$r$, and the policy predicts the two streams as $B=2$ successive blocks
(Sec.~\ref{sec:bicat_policy}).

\subsection{Bimanual Coordination Action Tokenization}
\label{sec:bicat_tokenization}

\paragraph{Canonical action space.}
To share motion representations across arms, we first align corresponding
action channels in a common coordinate convention. We retain the left-arm
coordinates and transform the right arm with an embodiment-specific sign
vector
$\mathbf{s}\in\{-1,+1\}^{d_a}$:
\begin{equation}
    \overline{\mathbf{A}}^{L}=\mathbf{A}^{L},
    \qquad
    \overline{\mathbf{A}}^{R}=\mathbf{A}^{R}\odot\mathbf{s},
    \label{eq:bicat_canonicalization}
\end{equation}
where the sign vector is broadcast over time. Each arm has its joint
coordinates followed by one gripper coordinate, and we use
$\mathbf{s}=(-1,1,1,1,-1,-1,1)$ for the RoboTwin robot and Bimanual YAM,
$\mathbf{s}=(1,-1,-1,1,-1,1,-1,1)$ for Unitree G1, and
$\mathbf{s}=(-1,1,1,1,-1,1)$ for Bimanual SO-101. 
Rather than normalizing each arm independently, we compute shared
channel-wise bounds $\boldsymbol{\ell},\mathbf{u}
\in\mathbb{R}^{d_a}$ by pooling corresponding action channels
across both canonical arms:
\begin{equation}
    \mathbf{X}^{r}
    =2\frac{\overline{\mathbf{A}}^{r}-\boldsymbol{\ell}}
    {\mathbf{u}-\boldsymbol{\ell}}-1,
    \qquad r\in\{L,R\}.
    \label{eq:bicat_normalization}
\end{equation}
All operations are element-wise, and the bounds are stored with the
tokenizer. This shared normalization ensures that corresponding
joint values are mapped identically across arms, aligning their
inputs before the shared encoder. We denote the sign transform followed by
shared normalization as the canonicalization $\mathcal{C}$, and its inverse
as $\mathcal{C}^{-1}$.

\begin{algorithm}[h]
\caption{BiCAT Tokenizer Training}
\label{alg:bicat_training}
\footnotesize
\begin{algorithmic}[1]
\Require Bimanual dataset $\mathcal{D}$;
canonicalization $\mathcal{C}$;
shared encoder $E_{\phi}$;
causal attention modules $\mathrm{SA},\mathrm{CA}$;
projection $P$; quantizer $\mathrm{FSQ}$;
joint decoder $D_{\theta}$.
\While{not converged}
    \State Sample $(\mathbf{A}^{L},\mathbf{A}^{R})
    \sim \mathcal{D}$
    \State \textbf{Canonicalization:}
    $(\mathbf{X}^{L},\mathbf{X}^{R})
    \gets \mathcal{C}(\mathbf{A}^{L},\mathbf{A}^{R})$
    \State \textbf{Encoding:}
    $\mathbf{H}^{L/R}
    \gets E_{\phi}(\mathbf{X}^{L/R})$
    \State \textbf{Coordination modeling:}
    $\mathbf{S}^{L/R}
    \gets \mathrm{SA}(\mathbf{H}^{L/R})$
    \Statex \hspace{\algorithmicindent}
    $\mathbf{C}^{L/R}
    \gets \mathbf{S}^{L/R}
    + \mathrm{CA}(\mathbf{S}^{L/R},
    \mathbf{S}^{R/L})$
    \State \textbf{Quantization:}
    $\hat{\mathbf{Z}}^{L/R}
    \gets \mathrm{FSQ}(P(\mathbf{C}^{L/R}))$
    \State \textbf{Temporal pairing:}
    $\mathbf{M}_i
    \gets \mathrm{Pair}(
    \hat{\mathbf{Z}}_i^{L},
    \hat{\mathbf{Z}}_i^{R})$
    \State \textbf{Joint decoding:}
    $(\hat{\mathbf{X}}^{L},\hat{\mathbf{X}}^{R})
    \gets D_{\theta}(\mathbf{M})$
    \State \textbf{Reconstruction loss:}
    $\mathcal{L}_{\mathrm{rec}}
    \gets \frac{1}{2Hd_a}
    \sum_{r\in\{L,R\}}
    \|\hat{\mathbf{X}}^{r}
    -\mathbf{X}^{r}\|_F^2$
    \State \textbf{Optimization:}
    $\Theta \gets \Theta
    -\eta\nabla_{\Theta}\mathcal{L}_{\mathrm{rec}}$
\EndWhile
\State $\mathcal{T}
\gets \{\mathcal{C},E_{\phi},
\mathrm{SA},\mathrm{CA},P,\mathrm{FSQ}\}$
\State $\mathcal{T}^{-1}
\gets \{\mathrm{Dequant},\mathrm{Pair},
D_{\theta},\mathcal{C}^{-1}\}$
\State \Return $\mathcal{T},\mathcal{T}^{-1}$
\end{algorithmic}
\end{algorithm}

\paragraph{Bimanual action encoding and quantization.}
We encode each arm's temporal structure with a shared encoder $E_{\phi}$,
a linear projection to width $d$ followed by two causal convolutional
stages:
\begin{equation}
    \mathbf{H}^{r}_{\ell}
    =\operatorname{Mish}\!\left(
    \operatorname{LN}\!\left(
    \operatorname{CConv}_{k_{\ell},s_{\ell}}
    (\mathbf{H}^{r}_{\ell-1})\right)\right),
    \quad \ell=1,2,
    \label{eq:bicat_temporal}
\end{equation}
where $\mathbf{H}^{r}_{0}$ is the projected action sequence. Kernels $(5,3)$
and strides $(2,2)$ compress $H=32$ steps into $M=8$ latent positions.
Each convolution uses left-only padding $k_{\ell}-s_{\ell}$, and layer
normalization acts on features independently at each timestep. Fixed
sinusoidal positions and shared pre-normalized causal self-attention produce
$\mathbf{S}^{L},\mathbf{S}^{R}\in\mathbb{R}^{M\times d}$.
Latent position $i$ thus depends only on actions up to timestep $4i$, using
one-based indexing. The projection, convolutions, normalization, and
self-attention parameters are shared across arms.

We next introduce partner-arm context through symmetric causal
cross-attention. Let $\operatorname{CA}(\mathbf{U},\mathbf{V})$ denote
multi-head attention~\cite{vaswani2017attention} with pre-normalized queries from $\mathbf{U}$ and
pre-normalized keys and values from $\mathbf{V}$. One shared attention block
updates both streams:
\begin{equation}
    \begin{aligned}
        \mathbf{C}^{L}&=\mathbf{S}^{L}
        +\operatorname{CA}(\mathbf{S}^{L},\mathbf{S}^{R}),\\
        \mathbf{C}^{R}&=\mathbf{S}^{R}
        +\operatorname{CA}(\mathbf{S}^{R},\mathbf{S}^{L}).
    \end{aligned}
    \label{eq:bicat_coordination}
\end{equation}
The residual update provides a direct path from each arm's own features while
adding context from its partner. Both directions read the original features
$(\mathbf{S}^{L},\mathbf{S}^{R})$, and query $i$ can attend only to partner
positions $j\leq i$. With dropout disabled, exchanging the
two canonical input streams exchanges their coordinated features. This
symmetry applies to the encoder; the two output streams retain explicit arm
indices and are each conditioned on the partner arm.

A shared linear projection maps each coordinated feature to
$\mathbf{v}^{r}_{i}\in\mathbb{R}^{d_z}$. Finite scalar quantization
(FSQ)~\cite{mentzer2024fsq} then bounds and rounds each coordinate to a fixed
grid, using a straight-through estimator during training. For scalar level
counts $(L_1,\ldots,L_{d_z})$, the grid digits
$n^{r}_{i,j}\in\{0,\ldots,L_j-1\}$ define a categorical symbol:
\begin{equation}
    y^{r}_{i}=\sum_{j=1}^{d_z}n^{r}_{i,j}\prod_{m<j}L_m,
    \qquad K=\prod_{j=1}^{d_z}L_j.
    \label{eq:bicat_quantization}
\end{equation}
We use $d_z=4$ and levels $(8,8,6,5)$, giving $K=1920$. Both arms share the
projection and grid, so the same symbol maps to the same quantized vector in
either stream. We serialize the tokens as
$\mathbf{y}=(y^{L}_{1},\ldots,y^{L}_{M},
y^{R}_{1},\ldots,y^{R}_{M})$, preserving two arm-indexed blocks from a
shared vocabulary.

\paragraph{Joint reconstruction and tokenizer training.}
Joint decoding reconstructs both trajectories from temporally aligned arm
codes. We dequantize each symbol into
$\widehat{\mathbf{z}}^{r}_{i}\in\mathbb{R}^{d_z}$ and concatenate the two
arm vectors at each latent position. A linear projection forms $M$ paired
memory slots:
\begin{equation}
    \mathbf{m}_{i}=\mathbf{W}_{m}
    [\widehat{\mathbf{z}}^{L}_{i}\Vert
    \widehat{\mathbf{z}}^{R}_{i}]+\mathbf{b}_{m}.
    \label{eq:bicat_paired_memory}
\end{equation}
A Transformer decoder $D_{\theta}$ reads this memory from $H$ zero-initialized queries
with fixed sinusoidal positions. A single output head predicts the normalized
bimanual chunk $\widehat{\mathbf{X}}\in\mathbb{R}^{H\times 2d_a}$.
Query self-attention is causal, while cross-attention accesses the complete
paired memory. All $H$ actions are reconstructed in parallel without previous
ground-truth actions as inputs. The decoder therefore reconstructs a complete
chunk from its latents, even though the encoder is causal. Inverse
normalization and the self-inverse right-arm sign transform return
$\widehat{\mathbf{A}}$ to the physical action convention.

We train the tokenizer solely through reconstruction in the shared normalized
space:
\begin{equation}
    \mathcal{L}_{\mathrm{rec}}
    =\mathbb{E}_{\mathbf{A}\sim\mathcal{D}_{\mathrm{valid}}}
    \left[\frac{\|\widehat{\mathbf{X}}-\mathbf{X}\|_{F}^{2}}
    {2Hd_a}\right],
    \label{eq:bicat_tokenizer_loss}
\end{equation}
where $\mathbf{X}=[\mathbf{X}^{L}\Vert\mathbf{X}^{R}]$ and
$\mathcal{D}_{\mathrm{valid}}$ contains chunks with no padded timesteps.
This objective jointly trains the temporal encoder, cross-arm attention,
quantization projection, and decoder, as summarized in
Algorithm~\ref{alg:bicat_training}. FSQ introduces no learnable codebook
or auxiliary commitment loss. We use width $d=256$, eight attention heads,
and dropout $0.1$. The decoder has six pre-normalized layers with feed-forward
width $4d$.

\subsection{Policy Training}
\label{sec:bicat_policy}

After tokenizer training, we freeze \textbf{BiCAT} and use its tokens to
supervise a pretrained VLM backbone with parameters $\theta_{\mathrm{V}}$:
PaliGemma~2-3B~\cite{steiner2024paligemma2} in simulation and
Molmo2-ER~\cite{fang2026molmoact2} on real robots (Sec.~\ref{sec:experiments}). The observation $\mathbf{o}$ forms the input prefix. We augment the vocabulary
with $K$ action symbols and an input-only mask token~\cite{chang2022maskgit}.

\paragraph{Semi-autoregressive action prediction.}
Following semi-autoregressive decoding~\cite{wang2018semiautoregressive},
the policy instantiates Eq.~(\ref{eq:ar_objective}) with the two arm streams
as successive blocks:
\begin{equation}
    p_{\theta_{\mathrm{V}}}(\mathbf{y}\mid\mathbf{o})
    =p_{\theta_{\mathrm{V}}}(\mathbf{y}^{L}\mid\mathbf{o})
    p_{\theta_{\mathrm{V}}}(\mathbf{y}^{R}\mid\mathbf{o},\mathbf{y}^{L}).
    \label{eq:bicat_bar}
\end{equation}
The $M=8$ tokens within each arm block are predicted in parallel, and the
right-arm block is conditioned on the complete left-arm block.
During training, the input suffix
$([\mathrm{MASK}]^{M},\mathbf{y}^{L})$ predicts
$(\mathbf{y}^{L},\mathbf{y}^{R})$ under block-causal attention. This lets the
right block use the complete ground-truth left block without exposing its
own targets. Both blocks receive equally weighted cross-entropy supervision
$\mathcal{L}_{\mathrm{CE}}$ in one forward pass. At inference, two sequential
block predictions generate the action tokens, and the frozen decoder
reconstructs the continuous chunk. We use greedy decoding restricted to
action symbols.

\section{Experiments}
\label{sec:experiments}

We evaluate \textbf{BiCAT} in simulation
(Secs.~\ref{sec:benchmark-and-baselines}--\ref{sec:simulation-control}) and on
real robots (Sec.~\ref{sec:real-robot-deployment}). Our experiments
assess both downstream policy performance and, through controlled
ablations, the impact of the key design choices of \textbf{BiCAT}.

\subsection{Simulation Setup}
\label{sec:benchmark-and-baselines}
\label{sec:training-protocol}

\paragraph{Benchmark and protocol.}
RoboTwin 2.0~\cite{chen2025robotwin2} comprises 50 bimanual manipulation
tasks. Tokenizers and policies are trained on all 2,500 demonstrations
from all 50 tasks. Each learned tokenizer is pretrained with its own
objective and then frozen, and a PaliGemma~2-3B
backbone~\cite{steiner2024paligemma2} is fully fine-tuned to predict its
action tokens from a language instruction, the discretized robot state, and
three RGB views. Policies predict 32-step chunks and execute the first 16
before replanning. As many of these tasks need only one arm, we
evaluate on ten representative tasks, selected before training
to cover three levels of inter-arm coupling, inspired by the coordination
taxonomy of \citet{sa2026vlareview}: two \emph{single-arm} tasks,
four \emph{loosely coupled} tasks where both arms act largely
independently, and four \emph{tightly coupled} tasks where each arm's
motion depends on its partner (Appendix~\ref{app:robotwin-tasks}). We
report success over 100 rollouts per task and the ten-task mean
(settings in Appendix~\ref{app:baselines}).

\paragraph{Baselines.}
We compare \textbf{BiCAT} with Bin~\cite{kim2024openvla},
FAST~\cite{pertsch2025fast}, QueST~\cite{mete2024quest},
OAT~\cite{liu2026oat_rss} with token-wise ($\mathrm{OAT}^{\mathrm{sing}}$)
and five-block ($\mathrm{OAT}^{\mathrm{pow2}}$) prediction,
$\mathcal{M}^{2}$Tok~\cite{xu2026m2tok}, NAC~\cite{jawaid2026nac}, and
ActionCodec~\cite{dong2026actioncodec}; Table~\ref{tab:efficiency} lists
their token budgets, vocabularies, sizes, prediction schedules, and
latencies. To keep the comparison controlled, all learned tokenizers use a
vocabulary of 1920 or 2048 codes, 16 tokens per chunk except NAC (24,
following its multi-scale design), and between 7.4M and 11.9M parameters,
with \textbf{BiCAT} the smallest. All tokenizers are fitted or trained on
the RoboTwin training data; FAST is refitted rather than taken from its
released universal tokenizer, and $\mathcal{M}^{2}$Tok uses its released
implementation (Appendix~\ref{app:tokenizer-configurations}). All methods
share the data, backbone,
observations, and optimization settings and differ only in the tokenizer
and its prediction schedule.

\begin{table}[!t]
    \centering
    \caption{Token budgets, prediction schedules, tokenizer size, and
    per-chunk latency on one H200 GPU. $K$: vocabulary size;
    $[b]\!\times\!n$: $n$ blocks of $b$ parallel tokens; Param.: tokenizer
    parameters (M), excluding NAC's training-only discriminator; Rec.:
    separately timed encode--decode round trip; Policy: mean token-generation
    time per chunk.}
    \label{tab:efficiency}
    \footnotesize
    \setlength{\tabcolsep}{3pt}
    \renewcommand{\arraystretch}{1.08}
    \fitwidth{%
    \begin{tabular}{@{}lcccccccc@{}}
        \toprule
        & & & & & \multicolumn{3}{c}{Tokenizer (ms)} & Policy \\
        \cmidrule(lr){6-8}
        Tokenizer & \#Tok. & $K$ & Schedule & Param. & Encode & Decode & Rec. & (ms) \\
        \midrule
        Bin & 448 & 256 & $[14]\!\times\!32$ & -- & 0.07 & 0.06 & 0.13 & 451.7 \\
        FAST & $N$ & 2048 & $[1]\!\times\!N$ & -- & 0.37 & 0.22 & 0.65 & 533.6 \\
        QueST & 16 & 1920 & $[1]\!\times\!16$ & 11.61 & 1.74 & 1.58 & 3.44 & 227.7 \\
        $\mathrm{OAT}^{\mathrm{sing}}$ & 16 & 1920 & $[1]\!\times\!16$ & 11.04 & 2.95 & 1.98 & 4.94 & 229.1 \\
        $\mathrm{OAT}^{\mathrm{pow2}}$ & 16 & 1920 & $[1,1,2,4,8]$ & 11.04 & 2.90 & 1.98 & 4.88 & 79.6 \\
        $\mathcal{M}^{2}$Tok & 16 & 2048 & $[8,8]$ & 11.27 & 2.05 & 1.98 & 3.92 & 40.6 \\
        NAC & 24 & 2048 & $[8,16]$ & 8.48 & 1.42 & 2.41 & 3.92 & 40.8 \\
        ActionCodec & 16 & 2048 & $[16]$ & 11.87 & 1.56 & 1.32 & 2.87 & 26.8 \\
        \midrule
        \textbf{BiCAT} & 16 & 1920 & $[8,8]$ & 7.40 & 1.74 & 1.64 & 3.45 & 40.5 \\
        \bottomrule
    \end{tabular}%
    }
    \vspace{-2mm}
\end{table}

\subsection{Simulation Results and Analysis}
\label{sec:simulation-control}

We use RoboTwin 2.0 to answer four research questions.
\textbf{(Q1)}~Does building bimanual structure into the tokenizer improve
bimanual VLA control over existing tokenizers?
\textbf{(Q2)}~Where does the improvement come from?
\textbf{(Q3)}~How sensitive is \textbf{BiCAT} to its vocabulary size, and
where does it fall short?
\textbf{(Q4)}~What does two-block prediction cost at inference time?

\begin{table*}[t]
    \centering
    \caption{
        Success rates on ten RoboTwin 2.0 tasks,
        evaluated over 100 rollouts per task and grouped by inter-arm
        coupling. Avg. denotes the unweighted task average.
        Best and second-best results are bolded
        and underlined, respectively.
    }
    \label{tab:main-tokenizer-results}

    \scriptsize
    \setlength{\tabcolsep}{3.2pt}
    \renewcommand{\arraystretch}{1.15}

    \resizebox{\textwidth}{!}{%
    \begin{tabular}{@{}l|ccccccccccc@{}}
        \toprule
        & \multicolumn{2}{c}{Single-arm}
        & \multicolumn{4}{c}{Loosely coupled}
        & \multicolumn{4}{c}{Tightly coupled} & \\
        \cmidrule(lr){2-3}\cmidrule(lr){4-7}\cmidrule(lr){8-11}

        \shortstack[c]{Tokenizer}
        &
        \shortstack[c]{Shake bottle\\horizontally}
        &
        \shortstack[c]{Place container\\plate}
        &
        \shortstack[c]{Stack blocks\\two}
        &
        \shortstack[c]{Dump bin\\bigbin}
        &
        \shortstack[c]{Stack bowls\\three}
        &
        \shortstack[c]{Pick dual\\bottles}
        &
        \shortstack[c]{Hand over\\mic}
        &
        \shortstack[c]{Place bread\\skillet}
        &
        \shortstack[c]{Place dual\\shoes}
        &
        \shortstack[c]{Put bottles\\dustbin}
        &
        \shortstack[c]{Avg.}
        \\
        
        \midrule

        Bin
        & 0.67 & 0.30 & 0.00 & 0.00 & 0.00 & 0.00 & 0.05 & 0.00 & 0.00 & 0.00 & 0.102 \\
        FAST
        & 0.70 & 0.25 & 0.15 & 0.27 & 0.09 & 0.11 & 0.52 & 0.13 & 0.01 & 0.00 & 0.223 \\
        QueST
        & 0.96 & \textbf{0.92} & 0.43 & 0.69 & 0.55 & 0.12 & \underline{0.90} & 0.10 & 0.04 & 0.20 & 0.491 \\
        $\mathrm{OAT}^{\mathrm{sing}}$
        & 0.93 & \underline{0.90} & 0.47 & 0.78 & 0.50 & 0.03 & \textbf{0.94} & 0.08 & 0.03 & 0.21 & 0.487 \\
        $\mathrm{OAT}^{\mathrm{pow2}}$
        & 0.93 & 0.85 & \underline{0.49} & 0.70 & 0.54 & 0.10 & 0.77 & 0.06 & 0.05 & 0.20 & 0.469 \\
        $\mathcal{M}^{2}$Tok
        & 0.87 & 0.69 & 0.18 & 0.53 & 0.43 & 0.12 & 0.83 & 0.14 & 0.05 & 0.21 & 0.405 \\
        NAC
        & \underline{0.99} & \underline{0.90} & 0.47 & \textbf{0.87} & \underline{0.57} & 0.15 & 0.74 & 0.18 & 0.04 & 0.26 & \underline{0.517} \\
        ActionCodec
        & 0.98 & 0.83 & 0.34 & 0.60 & 0.54 & \underline{0.23} & 0.72 & \underline{0.25} & \underline{0.06} & \underline{0.37} & 0.492 \\
        \oursrow\textbf{BiCAT (Ours)}
        & \textbf{1.00} & \textbf{0.92} & \textbf{0.66} & \underline{0.85} & \textbf{0.81} & \textbf{0.44} & 0.88 & \textbf{0.47} & \textbf{0.41} & \textbf{0.49} & \textbf{0.693} \\
        \bottomrule
    \end{tabular}%
    }

    \vspace{-2mm}
\end{table*}

\begin{figure*}[t]
    \centering
    \includegraphics[width=\textwidth]{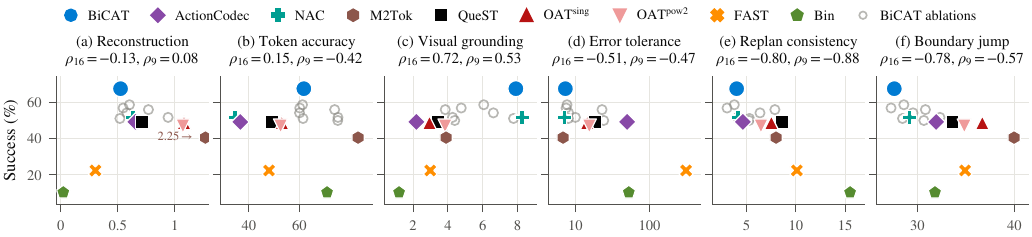}
    \caption{\textbf{What tracks closed-loop success.} Filled markers: nine
    tokenizers; hollow grey markers: seven BiCAT ablations
    (Table~\ref{tab:ablation}); y-axis: mean RoboTwin success. $\rho_{16}$ and
    $\rho_{9}$ are Spearman correlations over all 16 points and over the nine
    tokenizers.
    (a)~Reconstruction MSE of held-out chunks in the original action space
    ($\times10^{-3}$; $\mathcal{M}^{2}$Tok, 2.25, is clipped).
    (b)~Policy token accuracy (\%).
    (c)~Ratio of action-token NLL with shuffled versus matched images; higher
    means stronger reliance on vision.
    (d)~Change of the decoded action when $1/16$ of the tokens are replaced
    ($\times10^{-3}$, log scale).
    (e)~Disagreement between chunks predicted 8 steps apart over their 24
    overlapping steps ($\times10^{-3}$).
    (f)~Mean absolute change of the arm joints at chunk boundaries
    ($\times10^{-3}$\,rad). Lower is better in (d)--(f).}
    \label{fig:why-bicat-success}
\end{figure*}

\paragraph{Q1: How does BiCAT compare with baselines?}
As shown in Table~\ref{tab:main-tokenizer-results}, \textbf{BiCAT} reaches a mean
success rate of 69.3\%, compared with 51.7\% for NAC, the strongest
baseline. The gain of 17.6 percentage points
is accompanied by the best or tied-best result on eight of the ten tasks,
and it grows with inter-arm coupling: over the strongest baseline in each
group, \textbf{BiCAT} gains 1.5 points on the near-saturated single-arm
tasks, 17.5 on loosely coupled tasks, and 21.3 on tightly coupled tasks.
Since the learned tokenizers have comparable token budgets
(Sec.~\ref{sec:benchmark-and-baselines}), the gain is not explained by a shorter action sequence.
$\mathcal{M}^{2}$Tok, which also tokenizes each arm separately but with a
different architecture and multi-codebook quantization, reaches 40.5\%; we
isolate the role of cross-arm interaction with controlled ablations in Q2.

\paragraph{Q2: What explains the gains?}
Figure~\ref{fig:why-bicat-success} relates mean success to six properties of
each tokenizer and its trained policy. Consistent with
ActionCodec~\cite{dong2026actioncodec}, reconstruction fidelity does not
predict control ($\rho_{16}=-0.13$): Bin and FAST reconstruct actions most
accurately yet perform worst. Token accuracy is also uninformative
($\rho_{9}=-0.42$), as it is not comparable across vocabularies. Success
instead tracks reliance on visual input ($\rho_{16}=0.72$), consistency
across replanning ($\rho_{16}=-0.80$), and smoothness at chunk boundaries
($\rho_{16}=-0.78$), and less closely tolerance to token errors
($\rho_{16}=-0.51$); \textbf{BiCAT} has the smallest boundary jump among the
nine tokenizers. As eight of the 16 points are \textbf{BiCAT} or its
variants, we read these correlations as descriptive rather than causal.

\paragraph{Q2: Which components contribute?}
Table~\ref{tab:ablation} ablates the design principle and the prediction
schedule. \emph{Raw coordinates} sets $\mathbf{s}=\mathbf{1}$, so the two
arms share features for identical joint values rather than for mirrored
motions, and a motion such as reaching inward is encoded differently for
each arm; \emph{Arm-specific modules} gives each arm its own modules and
normalization bounds. These variants lower success to 58.7\% and 56.9\%;
sharing
aligns the latents of the two arms (left--right MMD of 0.088 for
\textbf{BiCAT}, versus 0.193 and 0.286; Appendix~\ref{app:arm-latents}). Removing the
pre-quantization cross-attention (\emph{Decoder-only coupling}), joint
decoding (\emph{Encoder-only coupling}), or both (\emph{Independent arms}, a
controlled per-arm variant) lowers success to
51.7\%, 56.0\%, and 50.0\%. The two interactions are complementary: each
adds far more when the other is present (17.6 versus 6.0 points for
cross-attention, 13.3 versus 1.7 for joint decoding), and removing them
weakens error tolerance and visual grounding. Predicting both arms in one
block (\emph{Single-block prediction}) lowers success to 51.1\%, so neither
partner-aware codes without the arm-wise schedule nor the schedule without
partner-aware codes exceeds NAC. Reversing the block order
(\emph{Right-first prediction}) lowers success to 54.2\%, so the order also
matters; we use left-first throughout.


\begin{table}[t]
    \centering
    \caption{Ablations on the ten RoboTwin 2.0 tasks. Columns mark the retained
    components: coordinate canonicalization, cross-arm sharing of modules and
    normalization, pre-quantization cross-attention, and joint decoding; Order
    is the prediction schedule (L$\to$R: left-arm block first; R$\to$L:
    right-arm block first; 1 blk: both arms in one block). SR is mean success (\%). Err.\ tol.\ is the change of the decoded
    action when $1/16$ of the tokens are replaced ($\times10^{-3}$, lower is
    better), and Vis.\ is the visual-grounding ratio of
    Fig.~\ref{fig:why-bicat-success}c (higher is better). Per-task results are
    in Appendix~\ref{app:ablation-per-task}.}
    \label{tab:ablation}
    \footnotesize
    \setlength{\tabcolsep}{2.6pt}
    \renewcommand{\arraystretch}{1.1}
    \newcommand{\yes}{$\checkmark$}
    \newcommand{\no}{--}
    \fitwidth{%
    \begin{tabular}{@{}lccccc|ccc@{}}
        \toprule
        Variant & Can. & Share & CA & Joint & Order & SR & Err.\ tol. & Vis. \\
        \midrule
        Raw coordinates & \no & \yes & \yes & \yes & L$\to$R & 58.7 & 7.7  & 6.1 \\
        Arm-specific modules & \yes & \no & \yes & \yes & L$\to$R & 56.9 & 7.5  & 4.8 \\
        \midrule
        Decoder-only coupling & \yes & \yes & \no & \yes & L$\to$R & 51.7 & 10.0 & 4.3 \\
        Encoder-only coupling & \yes & \yes & \yes & \no & L$\to$R & 56.0 & 22.9 & 3.8 \\
        Independent arms & \yes & \yes & \no & \no & L$\to$R & 50.0 & 24.1 & 4.4 \\
        \midrule
        Single-block prediction & \yes & \yes & \yes & \yes & 1 blk & 51.1 & 8.1  & 7.8 \\
        Right-first prediction   & \yes & \yes & \yes & \yes & R$\to$L & 54.2 & 8.1 & 6.6 \\
        \midrule
        \textbf{BiCAT} & \yes & \yes & \yes & \yes & L$\to$R & \textbf{69.3} & \textbf{7.3} & \textbf{7.9} \\
        \bottomrule
    \end{tabular}%
    }
    \vspace{-2mm}
\end{table}

\begin{table}[!htbp]
    \centering
    \caption{FSQ grid ablation. Mean success rate over the ten RoboTwin 2.0
    tasks for different FSQ levels and their induced vocabulary sizes.}
    \label{tab:FSQ}
    \footnotesize
    \setlength{\tabcolsep}{3.5pt}
    \renewcommand{\arraystretch}{1.1}
    \fitwidth{%
    \begin{tabular}{@{}l|ccccc@{}}
        \toprule
        FSQ Levels
        & [8,6,5]
        & [8,8,8]
        & [8,5,5,5]
        & [8,8,6,5]
        & [7,5,5,5,5] \\
        Induced $|\mathcal{V}|$
        & 240 & 512 & 1000 & 1920 & 4375 \\
        \midrule
        Mean SR
        & 0.470
        & 0.617
        & 0.631
        & 0.693
        & 0.589 \\
        \bottomrule
    \end{tabular}%
    }
\end{table}

\paragraph{Q3: How does the vocabulary affect control?}
Table~\ref{tab:FSQ} varies the FSQ levels of \textbf{BiCAT} (per-task results
in Appendix~\ref{app:fsq-per-task}). We choose the grids following
FSQ~\cite{mentzer2024fsq}: all levels satisfy $L_j\geq5$, and $[8,6,5]$,
$[8,5,5,5]$, and $[7,5,5,5,5]$ are its recommended configurations for
vocabularies of about $2^{8}$, $2^{10}$, and $2^{12}$. Success rises from
47.0\% with 240 codes to 69.3\% with 1920 codes and falls to 58.9\% with
4375 codes, so additional quantization capacity does not monotonically
improve control, although changing the levels also changes the latent
dimensionality. All other experiments use $[8,8,6,5]$, as do QueST and OAT.

\paragraph{Q3: Where does BiCAT underperform?}
The improvement is not uniform. \textbf{BiCAT} reaches 88\% on
\textsc{Handover Mic}, below $\mathrm{OAT}^{\mathrm{sing}}$ (94\%) and QueST
(90\%), and 85\% on \textsc{Dump Bin Bigbin}, below NAC (87\%). On
\textsc{Handover Mic}, four tokenizer ablations with the same block order
reach 92--96\% (Appendix~\ref{app:ablation-per-task}), so this gap is not
inherent to arm-wise tokens or two-block prediction.

\paragraph{Q4: What does two-block prediction cost?}
Table~\ref{tab:efficiency} reports per-chunk latency on one H200 GPU with
batch size 1. Encoding and decoding take under 5\,ms for every tokenizer,
so latency is set by the prediction schedule. With two blocks, \textbf{BiCAT} takes 40.5\,ms, comparable to
$\mathcal{M}^{2}$Tok and NAC, which also predict two blocks, and
$5.6\times$ to over $11\times$ faster than tokenizers decoded token by token.
Only ActionCodec, which predicts a single block, is faster (26.8\,ms), but
\emph{Single-block prediction} costs \textbf{BiCAT} 18.2 points of success,
so the second block is worth its extra decoding step.

\subsection{Deployment on Real Robots}
\label{sec:real-robot-deployment}

\begin{figure*}[!htbp]
    \centering
    \includegraphics[width=\textwidth]{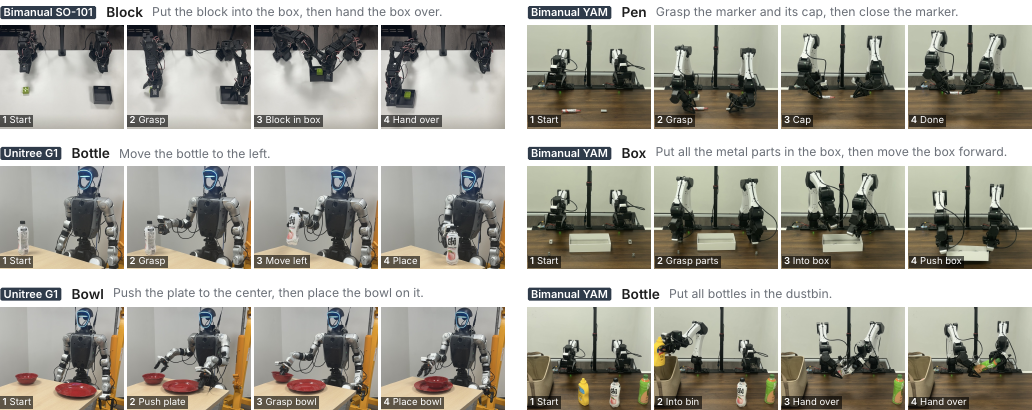}
    \caption{\textbf{BiCAT rollouts on three real bimanual embodiments.}
    Each row shows four keyframes of one task with a short task description.}
    \label{fig:real-rollouts}
\end{figure*}

\paragraph{Platforms and tasks.}
We evaluate on three bimanual embodiments: Bimanual SO-101, the Unitree G1
humanoid, and Bimanual YAM. The suite contains six tasks: Block on SO-101,
Bottle and Bowl on G1, and Pen, Box, and Bottle on YAM.
Fig.~\ref{fig:real-rollouts} shows each task, and
Appendix~\ref{app:real-world} gives task statistics.

\begin{table*}[!htbp]
    \centering
    \caption{Real-world task success and inference latency, grouped by embodiment.
    Success is reported as successful trials over evaluated trials.
    Overall pools trials across all three embodiments, and latency reports
    the YAM measurements.}
    \label{tab:real-world-results}
    \label{tab:three-task-success-counts}
    \label{tab:unitree-task-success-counts}
    \label{tab:so101-task-success-counts}
    \footnotesize
    \setlength{\tabcolsep}{3.5pt}
    \renewcommand{\arraystretch}{1.15}
    \begin{tabularx}{\textwidth}{@{}l*{7}{>{\centering\arraybackslash}X}>{\centering\arraybackslash}p{2.0cm}@{}}
        \toprule
        & \textbf{SO-101}
        & \multicolumn{2}{c}{\textbf{Unitree G1}}
        & \multicolumn{3}{c}{\textbf{YAM}}
        & \textbf{Overall} & \textbf{Latency} \\
        \cmidrule(lr){2-2}
        \cmidrule(lr){3-4}
        \cmidrule(lr){5-7}
        \cmidrule(lr){8-8}
        \cmidrule(l){9-9}
        Tokenizer & Block & Bottle & Bowl & Pen & Box & Bottle & All tasks & (ms) \\
        \midrule
        FAST~\cite{pertsch2025fast}
        & 3/25 & 6/25 & 7/25 & 0/25 & 1/25 & 8/25 & 25/150 & $434.1 \pm 103.6$ \\
        $\mathrm{OAT}^{\mathrm{sing}}$~\cite{liu2026oat_rss}
        & 19/25 & 13/25 & 15/25 & 5/25 & 7/25 & 15/25 & 74/150 & $251.6 \pm 5.1$ \\
        \oursrow\textbf{BiCAT (Ours)}
        & \textbf{25/25} & \textbf{24/25} & \textbf{21/25} & \textbf{8/25} & \textbf{14/25} & \textbf{21/25} & \textbf{113/150} & \textbf{91.7 $\pm$ 1.3} \\
        \bottomrule
    \end{tabularx}
\end{table*}

\paragraph{Data and training.}
We collect 100 teleoperated demonstrations per task (105 for Pen). For each
embodiment, one tokenizer is trained on all of its demonstrations and
frozen, and one VLA policy with a Molmo2-ER
backbone~\cite{fang2026molmoact2} is trained on its tokens for all tasks of
that embodiment. Within an embodiment, all methods use exactly the same
data and training configuration, including training steps, learning rate,
warmup, and batch size (Appendix~\ref{app:real-world-training}), and the
final checkpoint is evaluated.

\paragraph{Baselines and evaluation.}
We compare \textbf{BiCAT} with one representative tokenizer from each of the
two main families: FAST~\cite{pertsch2025fast}, the widely adopted
compression-based tokenizer, and $\mathrm{OAT}^{\mathrm{sing}}$~\cite{liu2026oat_rss},
a learned tokenizer with ordered tokens and token-wise prediction. All
policies run on one NVIDIA RTX 5090 GPU and are evaluated for 25 trials per
task, 150 per method. A trial succeeds only if every stage of the task is
completed, and the initial object placements are identical across methods.
Inference latency is measured on YAM, from receiving an observation to
outputting an action chunk, over all inference calls.
The robot waits during inference, so latency lengthens an episode but does
not change the observations on which chunks are predicted
(Appendix~\ref{app:real-world-eval}).

\paragraph{Results.}
As shown in Table~\ref{tab:real-world-results}, \textbf{BiCAT} succeeds in 113
of 150 trials, compared with 74 for $\mathrm{OAT}^{\mathrm{sing}}$ and 25 for
FAST, leads on all six tasks, and has the lowest inference latency, $4.7\times$
lower than FAST. Consistent with simulation, the advantage persists across physical bimanual embodiments.



\section{Conclusion \& Limitations}
\label{sec:conclusion}
We introduced \textbf{BiCAT}, a bimanual action tokenizer for autoregressive VLAs that combines a shared motion representation across mirrored arms with partner-conditioned discretization, while preserving separate token streams for each arm. The policy predicts these streams in two successive blocks, requiring only two decoding steps per action chunk. Across simulation and three real-world bimanual embodiments, \textbf{BiCAT} achieves the highest success rates among the evaluated tokenizers, with larger gains on tasks requiring tighter coordination. Ablations show that both cross-arm interaction within the tokenizer and blockwise prediction in the policy are necessary for these gains over joint tokenization, highlighting action tokenization as an effective way to model bimanual coordination. However, \textbf{BiCAT} currently operates in joint space: its fixed per-joint sign mapping does not directly extend to end-effector actions, which require different transformations for positions and orientations. Each tokenizer is also trained for a single embodiment and assigns equal, fixed token budgets to both arms. Extending \textbf{BiCAT} to end-effector actions, cross-embodiment transfer, and adaptive token allocation remains future work.


\bibliographystyle{abbrvnat}
\bibliography{reference}

\clearpage
\beginappendix

\begin{figure}[!ht]
    \centering
    \includegraphics[width=\textwidth]{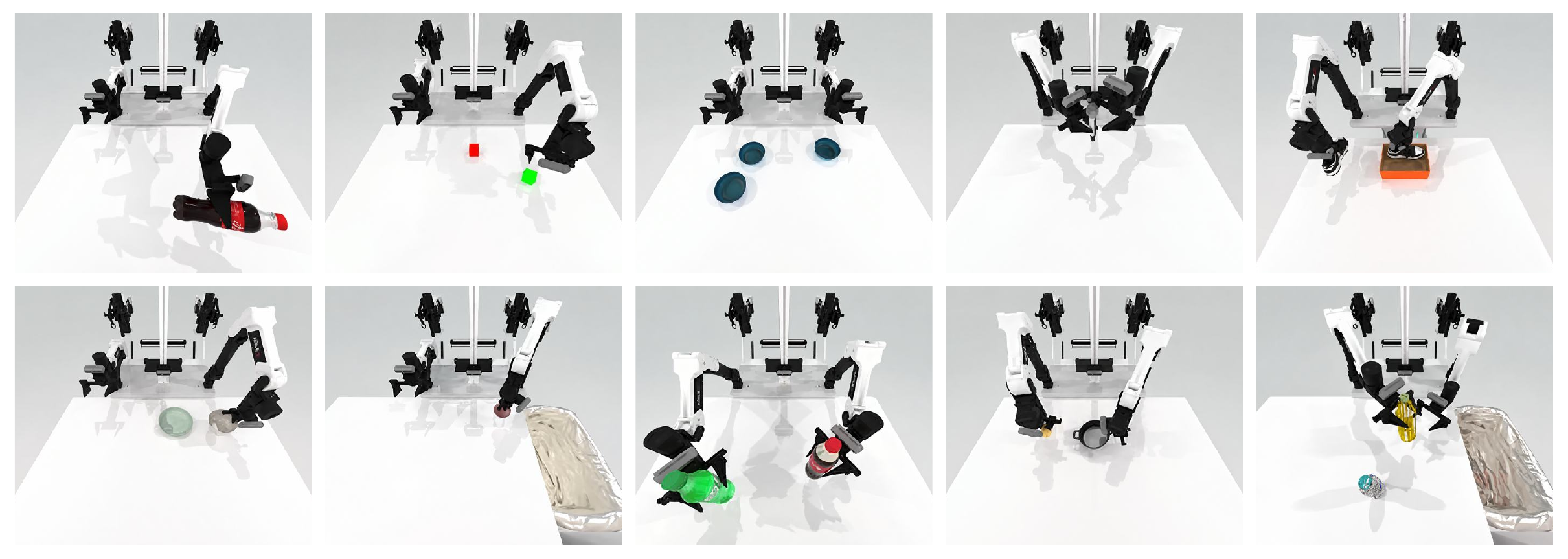}
    \caption{Overview of the ten RoboTwin evaluation tasks.
    Task names and instructions are listed in
    Table~\ref{tab:robotwin-task-details}.}
    \label{fig:supp-simulation-task-overview}
\end{figure}

\section{Simulation Task Overview}
\label{app:robotwin-tasks}

We evaluate on ten tasks from RoboTwin 2.0~\cite{chen2025robotwin2},
illustrated in Fig.~\ref{fig:supp-simulation-task-overview}.
Table~\ref{tab:robotwin-task-details} lists their names, representative
instructions, and average episode lengths.

\begin{table}[!ht]
    \centering
    \caption{Overview of all ten RoboTwin evaluation tasks. Avg. steps
    denotes average recorded demonstration length, not the rollout limit.
    Instructions summarize the task goals.}
    \label{tab:robotwin-task-details}
    \footnotesize
    \setlength{\tabcolsep}{2.5pt}
    \renewcommand{\arraystretch}{1.12}
    \begin{tabular}{@{}
        >{\raggedright\arraybackslash}p{0.31\columnwidth}
        >{\raggedleft\arraybackslash}p{0.085\columnwidth}
        >{\raggedright\arraybackslash}p{\dimexpr0.605\columnwidth-4\tabcolsep\relax}
        @{}}
        \toprule
        Task & \shortstack{Avg.\\steps} & Instruction \\
        \midrule
        Shake Bottle Horizontally & 276
        & Pick up the bottle and shake it horizontally. \\
        Place Container Plate & 156
        & Place the container on the plate. \\
        Stack Blocks Two & 316
        & Move the red block to the center and stack the green block on it. \\
        Dump Bin Bigbin & 265
        & Pour the balls from the small bin into the large bin. \\
        Stack Bowls Three & 476
        & Stack the three bowls on top of one another. \\
        Pick Dual Bottles & 127
        & Pick up one bottle with each arm simultaneously. \\
        Handover Mic & 223
        & Pick up the microphone and hand it to the other arm. \\
        Place Bread Skillet & 162
        & Hold the skillet with one arm and place the bread inside with the other. \\
        Place Dual Shoes & 228
        & Place both shoes in the shoebox with their tips pointing left. \\
        Put Bottles Dustbin & 637
        & Put all three bottles into the dustbin on the left. \\
        \bottomrule
    \end{tabular}
\end{table}


\subsection{Evaluation Protocol}
\label{app:robotwin-eval-protocol}

Each method is evaluated with 100 rollouts per task using different random
seeds. We use RoboTwin's task-specific success checks and report both
per-task success and the unweighted mean across the ten tasks.
Policies predict 32-step action chunks, execute the first 16 steps, and
then replan from a new observation. All ten evaluation tasks belong to
the 50-task collection used for training.

\section{Implementation Details}
\label{app:baselines}

This section complements Sec.~\ref{sec:benchmark-and-baselines} with the
simulation training details of BiCAT and the baselines.

\subsection{Training Data and Inputs}
\label{app:baseline-data}

The 5\% validation split is drawn at the episode level with a fixed seed,
giving 2,375 training and 125 validation episodes, and is shared by
tokenizer and policy training. Action chunks of 32 steps are sampled within
episode boundaries; each step has 14 dimensions, six joint values and one
gripper value per arm. The three RGB views come from the head and the two
wrist cameras, and the robot state is encoded as a text prefix with 256
bins per dimension. All policies use the
\texttt{paligemma2-3b-mix-224} checkpoint.

\subsection{Optimization Settings}
\label{app:baseline-protocol}

Table~\ref{tab:supp-training-settings} lists the optimization settings
shared by all methods. Policies minimize teacher-forced action-token
cross-entropy and use greedy decoding during evaluation.

\begin{table}[!ht]
    \centering
    \caption{Shared optimization settings for the simulation comparison.
    Tokenizer pretraining applies to QueST, both OAT variants,
    $\mathcal{M}^{2}$Tok, NAC, ActionCodec, and BiCAT. The policy learning rate applies equally to
    the language model, vision encoder, and modality bridge.}
    \label{tab:supp-training-settings}
    \footnotesize
    \setlength{\tabcolsep}{3pt}
    \renewcommand{\arraystretch}{1.15}
    \begin{tabular}{@{}
        >{\raggedright\arraybackslash}p{0.40\columnwidth}
        >{\centering\arraybackslash}p{0.30\columnwidth}
        >{\centering\arraybackslash}p{\dimexpr0.30\columnwidth-4\tabcolsep\relax}
        @{}}
        \toprule
        Setting & \shortstack{Tokenizer\\pretraining}
        & \shortstack{Policy\\training} \\
        \midrule
        Optimizer steps & 300k & 30k \\
        Batch size & 512 & 64 \\
        Optimizer & AdamW & AdamW \\
        Learning rate & $5\times10^{-5}$ & $10^{-4}$ \\
        Adam $\boldsymbol{\beta}$ & $(0.9,0.95)$ & $(0.95,0.999)$ \\
        Adam $\epsilon$ & $10^{-8}$ & $10^{-6}$ \\
        Weight decay & 0 & $10^{-6}$ \\
        Warmup steps & 200 & 500 \\
        Schedule after warmup & Constant & Cosine \\
        Final learning rate & $5\times10^{-5}$ & $10^{-5}$ \\
        Precision & BF16 & BF16 \\
        Gradient norm limit & 1.0 & 1.0 \\
        Validation interval & 1k steps & 1k steps \\
        \bottomrule
    \end{tabular}
\end{table}

\subsection{Tokenizer Configurations}
\label{app:tokenizer-configurations}

Table~\ref{tab:efficiency} in the main paper lists the token count,
vocabulary size, and prediction schedule of each tokenizer. All learned
tokenizers use a hidden width of 256 unless noted and are trained on action
reconstruction.
\begin{itemize}
    \item \textbf{Bin}~\cite{kim2024openvla}: each normalized action value is
    mapped to one of 256 uniform bins.
    \item \textbf{FAST}~\cite{pertsch2025fast}: a discrete cosine transform,
    coefficient quantization, and byte-pair encoding, fitted on the RoboTwin
    training split rather than taken from the released universal tokenizer;
    it yields 26.8 tokens per chunk on average (range 9--80).
    \item \textbf{QueST}~\cite{mete2024quest}: causal temporal convolutions
    followed by a six-layer causal Transformer encoder, with FSQ levels
    $[8,8,6,5]$.
    \item \textbf{OAT}~\cite{liu2026oat_rss}: learned register tokens trained
    with nested prefix dropout over $\{1,2,4,8,16\}$ registers, with FSQ
    levels $[8,8,6,5]$; the two variants differ only in their register and
    prediction block schedules.
    \item $\boldsymbol{\mathcal{M}^{2}}$\textbf{Tok}~\cite{xu2026m2tok}: the
    released implementation with width 120. A shared convolutional VQ-VAE
    encodes each arm independently into two latent frames, each quantized by
    four cosine-similarity codebooks of 512 entries, and reconstructs each
    arm from its own tokens.
    \item \textbf{NAC}~\cite{jawaid2026nac}: a SEANet encoder over the
    time-major flattened chunk and two-level residual FSQ with levels
    $[8,8,4,4]$, a coarse level of 8 tokens and a fine level of 16; its
    Vocos decoder is trained with reconstruction, adversarial, and
    feature-matching losses.
    \item \textbf{ActionCodec}~\cite{dong2026actioncodec}: a six-layer
    Perceiver-style encoder and decoder, with 16 latent vectors quantized by
    a 2048-entry VQ codebook updated by exponential moving average.
    \item \textbf{BiCAT}: the architecture of Sec.~\ref{sec:method}, with one
    shared causal self-attention layer and one symmetric cross-arm attention
    layer in the encoder, FSQ levels $[8,8,6,5]$, and a six-layer joint
    decoder.
\end{itemize}

\section{Per-Task Ablation Results}
\label{app:ablation-per-task}

Table~\ref{tab:supp-ablation-per-task} reports per-task success for the
variants summarized in Table~\ref{tab:ablation}.

\section{Shared Arm Representations}
\label{app:arm-latents}

\begin{figure*}[!ht]
    \centering
    \newcommand{\armpanel}[3]{%
        \begin{minipage}[t]{0.32\textwidth}\centering
        {\small #1}\\[2pt]
        \includegraphics[width=\linewidth,trim=#2 53.6 #3 27,clip]{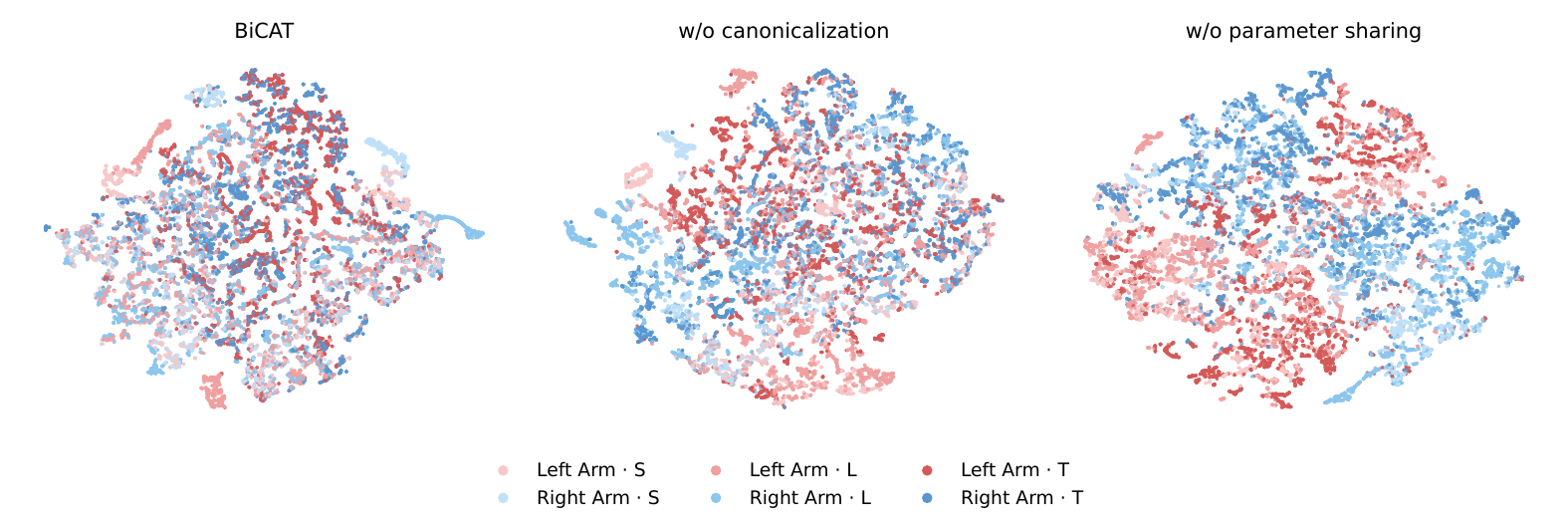}
        \end{minipage}}
    \armpanel{BiCAT}{18}{519.7}\hfill
    \armpanel{\emph{Raw coordinates}}{268.4}{269.2}\hfill
    \armpanel{\emph{Arm-specific modules}}{519}{18.6}\\[3pt]
    {\small\textcolor[RGB]{208,88,88}{\large$\bullet$}~Left arm\qquad
    \textcolor[RGB]{88,144,208}{\large$\bullet$}~Right arm}
    \caption{t-SNE of pre-quantization arm latents on the ten RoboTwin 2.0
    tasks. Each point is one arm of one action chunk, averaged over its eight
    latent positions; each panel contains the same 10,000 chunks. Red and
    blue shades denote the left and right arms. Each panel is fitted
    separately, so coordinates and distances are not comparable across
    panels; the left--right MMD reported in the text is computed in the
    original latent space.}
    \label{fig:arm-latents}
\end{figure*}

Figure~\ref{fig:arm-latents} visualizes the pre-quantization latents of
BiCAT, \emph{Raw coordinates}, and \emph{Arm-specific modules}. We encode
10,000 action chunks from the ten evaluation tasks, average each arm's eight
pre-quantization latent vectors, and embed the left- and right-arm averages
of each model jointly with t-SNE. With all modules shared, the latents of
the two arms overlap. With \emph{Arm-specific modules}, the two arms occupy
largely separate regions, so the shared FSQ grid assigns the same token IDs
to latents that are distributed differently for the two arms.

We quantify the overlap with the maximum mean discrepancy (MMD) between the
left- and right-arm averages in the original 4-D latent space, using an RBF
kernel whose bandwidth is the median pairwise squared distance. The MMD is
0.088 for BiCAT, 0.193 for \emph{Raw coordinates}, and 0.286 for
\emph{Arm-specific modules}, and the ordering is unchanged after pooled
per-dimension standardization. Under \emph{Raw coordinates}, the two arms
remain visually mixed in the t-SNE embedding, yet their MMD is more than
twice that of BiCAT, so canonicalization aligns the arms beyond what the
shared encoder alone achieves. These distances describe arm-marginal
distributions: they are consistent with, but do not directly measure, a
token ID denoting the same motion for both arms, and they do not by
themselves measure coordination or task success.

\section{Per-Task FSQ Ablation}
\label{app:fsq-per-task}

Table~\ref{tab:supp-fsq-per-task} reports per-task success for the FSQ
grids summarized in Table~\ref{tab:FSQ}. All grids use the BiCAT
architecture and the common training and evaluation protocol.

\section{Real-World Experiment Details}
\label{app:real-world}

Table~\ref{tab:real-world-task-details} lists the six real-world tasks.
All demonstrations are collected by teleoperation and recorded at 30\,Hz,
which is also the control frequency at deployment.

\subsection{Training Settings}
\label{app:real-world-training}

\begin{table}[!ht]
    \centering
    \caption{Real-world optimizer steps per embodiment. All tokenizers on
    the same embodiment use the same numbers of steps.}
    \label{tab:real-world-training}
    \footnotesize
    \setlength{\tabcolsep}{6pt}
    \renewcommand{\arraystretch}{1.12}
    \begin{tabular}{@{}lcc@{}}
        \toprule
        Embodiment & Tokenizer steps & Policy steps \\
        \midrule
        Bimanual SO-101 & 50k  & 4k \\
        Unitree G1     & 80k  & 9k \\
        Bimanual YAM   & 300k & 20k \\
        \bottomrule
    \end{tabular}
\end{table}

Each embodiment is trained separately. For each embodiment, one action
tokenizer is trained on the demonstrations of all its tasks in
Table~\ref{tab:real-world-task-details} without a held-out validation
split and then frozen, and one VLA policy is trained on its
tokens jointly for all of these tasks; on Unitree G1, for example, a single
tokenizer and a single policy cover both Bottle and Bowl.
Table~\ref{tab:real-world-training} lists the numbers of optimizer steps,
which scale with the amount of data on each embodiment. BiCAT and
$\mathrm{OAT}^{\mathrm{sing}}$ use the same tokenizer architectures as in
simulation, with 16 tokens per chunk and FSQ levels $[8,8,6,5]$, and FAST is
fitted separately on the demonstrations of each embodiment.

\paragraph{Policy.}
The real-world policies use Molmo2-ER~\cite{fang2026molmoact2} as the
vision-language backbone, which is fully fine-tuned together with the
vision encoder and the modality bridge. The policy observes the language
instruction, the joint state of both arms, and three RGB views from the
current time step: a top camera and the two wrist cameras. On Bimanual YAM
and Unitree G1 the top view has a resolution of $360\times640$ and the
wrist views $480\times640$; on Bimanual SO-101 all three views are
$480\times640$. The policy predicts 32-step action chunks as in
simulation, but on the real robots it executes all 32 steps of each chunk
before replanning, rather than the first 16.

\paragraph{Optimization.}
Tokenizer pretraining uses the settings of
Table~\ref{tab:supp-training-settings} except for the number of steps
(Table~\ref{tab:real-world-training}). Policy training uses the same
peak learning rate and cosine decay as in simulation, with a global batch
size of 128, BF16 mixed precision, and gradient-norm clipping at 1.0.
Within an embodiment, the training configuration is identical across
methods: BiCAT and $\mathrm{OAT}^{\mathrm{sing}}$ are trained for the same
number of tokenizer steps, FAST is fitted on the same demonstrations, and
all three policies share the data, observations, number of policy steps,
learning rate, warmup, and batch size. They differ only in the tokenizer
and its prediction schedule.

\subsection{Evaluation Protocol}
\label{app:real-world-eval}

Each method is evaluated for 25 trials per task. A trial counts as a
success only if the robot completes every stage of the task without error;
on Block, for example, the block must be placed in the box and the box
handed over. Partial completion counts as a failure. During deployment, the
robot waits while the policy predicts the next action chunk and executes
the chunk once inference has finished, so differences in inference latency
lengthen the episode but do not change the observation on which each chunk
is predicted. The initial object placements are identical across methods
for each trial, and each trial is limited to 120\,s, after which it counts
as a failure. Inference latency is measured from receiving an observation to outputting
an action chunk, over all inference calls during the YAM evaluation, and is
reported as mean $\pm$ standard deviation.

\begin{table*}[!ht]
    \centering
    \caption{Per-task success rates of the BiCAT ablations on the ten
    RoboTwin 2.0 tasks, evaluated over 100 rollouts per task.}
    \label{tab:supp-ablation-per-task}
    \footnotesize
    \setlength{\tabcolsep}{3.5pt}
    \resizebox{\textwidth}{!}{%
    \begin{tabular}{@{}l*{10}{c}c@{}}
        \toprule
        Variant
        & \shortstack[c]{Shake bottle\\horizontally}
        & \shortstack[c]{Place container\\plate}
        & \shortstack[c]{Stack blocks\\two}
        & \shortstack[c]{Dump bin\\bigbin}
        & \shortstack[c]{Stack bowls\\three}
        & \shortstack[c]{Pick dual\\bottles}
        & \shortstack[c]{Hand over\\mic}
        & \shortstack[c]{Place bread\\skillet}
        & \shortstack[c]{Place dual\\shoes}
        & \shortstack[c]{Put bottles\\dustbin}
        & Avg. \\
        \midrule
        Raw coordinates & 0.93 & 0.88 & 0.66 & 0.85 & 0.64 & 0.22 & 0.95 & 0.26 & 0.24 & 0.24 & 0.587 \\
        Arm-specific modules & 0.94 & 0.87 & 0.51 & 0.85 & 0.75 & 0.21 & 0.94 & 0.24 & 0.05 & 0.33 & 0.569 \\
        \midrule
        Decoder-only coupling & 0.95 & 0.73 & 0.44 & 0.68 & 0.55 & 0.12 & 0.96 & 0.31 & 0.10 & 0.33 & 0.517 \\
        Encoder-only coupling & 0.96 & 0.76 & 0.47 & 0.62 & 0.65 & 0.31 & 0.92 & 0.42 & 0.29 & 0.20 & 0.560 \\
        Independent arms & 0.89 & 0.82 & 0.37 & 0.73 & 0.58 & 0.34 & 0.70 & 0.38 & 0.08 & 0.11 & 0.500 \\
        \midrule
        Single-block prediction & 0.90 & 0.89 & 0.61 & 0.68 & 0.64 & 0.17 & 0.62 & 0.21 & 0.06 & 0.33 & 0.511 \\
        Right-first prediction & 0.97 & 0.87 & 0.47 & 0.84 & 0.61 & 0.17 & 0.92 & 0.37 & 0.04 & 0.16 & 0.542 \\
        \midrule
        \oursrow\textbf{BiCAT} & 1.00 & 0.92 & 0.66 & 0.85 & 0.81 & 0.44 & 0.88 & 0.47 & 0.41 & 0.49 & 0.693 \\
        \bottomrule
    \end{tabular}%
    }
\end{table*}

\begin{table*}[!ht]
    \centering
    \caption{Per-task success rates of BiCAT with different FSQ levels on
    the ten RoboTwin 2.0 tasks, evaluated over 100 rollouts per task.
    $|\mathcal{V}|$ is the induced vocabulary size, and Avg.\ is the
    unweighted task average.}
    \label{tab:supp-fsq-per-task}
    \footnotesize
    \setlength{\tabcolsep}{3.5pt}
    \resizebox{\textwidth}{!}{%
    \begin{tabular}{@{}lc*{10}{c}c@{}}
        \toprule
        FSQ levels & $|\mathcal{V}|$
        & \shortstack[c]{Shake bottle\\horizontally}
        & \shortstack[c]{Place container\\plate}
        & \shortstack[c]{Stack blocks\\two}
        & \shortstack[c]{Dump bin\\bigbin}
        & \shortstack[c]{Stack bowls\\three}
        & \shortstack[c]{Pick dual\\bottles}
        & \shortstack[c]{Hand over\\mic}
        & \shortstack[c]{Place bread\\skillet}
        & \shortstack[c]{Place dual\\shoes}
        & \shortstack[c]{Put bottles\\dustbin}
        & Avg. \\
        \midrule
        $[8,6,5]$     & 240  & 0.87 & 0.86 & 0.47 & 0.69 & 0.51 & 0.12 & 0.82 & 0.12 & 0.04 & 0.20 & 0.470 \\
        $[8,8,8]$     & 512  & 0.99 & 0.84 & 0.65 & 0.84 & 0.70 & 0.22 & 0.97 & 0.35 & 0.21 & 0.40 & 0.617 \\
        $[8,5,5,5]$   & 1000 & 0.92 & 0.91 & 0.61 & 0.84 & 0.71 & 0.21 & 0.97 & 0.42 & 0.21 & 0.51 & 0.631 \\
        $[8,8,6,5]$   & 1920 & 1.00 & 0.92 & 0.66 & 0.85 & 0.81 & 0.44 & 0.88 & 0.47 & 0.41 & 0.49 & 0.693 \\
        $[7,5,5,5,5]$ & 4375 & 0.90 & 0.85 & 0.51 & 0.81 & 0.61 & 0.34 & 0.91 & 0.34 & 0.23 & 0.39 & 0.589 \\
        \bottomrule
    \end{tabular}%
    }
\end{table*}

\begin{table*}[!ht]
    \centering
    \caption{Real-world tasks. The task description summarizes each task and
    need not match the language instruction given to the policy. Action dim.\
    is the per-arm action dimension times the number of arms. Demos is the
    number of teleoperated demonstrations, and Avg.\ steps is the average
    number of recorded time steps per demonstration.}
    \label{tab:real-world-task-details}
    \footnotesize
    \setlength{\tabcolsep}{4pt}
    \renewcommand{\arraystretch}{1.12}
    \begin{tabular}{@{}llp{0.42\textwidth}cccc@{}}
        \toprule
        Embodiment & Task & Task description & Action dim. & Demos & Avg.\ steps & Rec.\ rate \\
        \midrule
        Bimanual SO-101 & Block  & Put the block into the box, then hand the box over. & $6\times2$ & 100 & 544 & 30\,Hz \\
        \midrule
        Unitree G1 & Bottle & Move the bottle to the left. & $8\times2$ & 100 & 441 & 30\,Hz \\
                   & Bowl   & Push the plate to the center, then place the bowl on it. & $8\times2$ & 100 & 538 & 30\,Hz \\
        \midrule
        Bimanual YAM & Pen    & Grasp the marker and its cap, then close the marker. & $7\times2$ & 105 & 447 & 30\,Hz \\
                     & Box    & Put all the metal parts in the box, then move the box forward. & $7\times2$ & 100 & 733 & 30\,Hz \\
                     & Bottle & Put all bottles in the dustbin. & $7\times2$ & 100 & 1236 & 30\,Hz \\
        \bottomrule
    \end{tabular}
\end{table*}

\end{document}